\documentclass[10pt,twocolumn,letterpaper]{article}
\usepackage[pagenumbers]{cvpr} 

\usepackage{graphicx}
\usepackage{colortbl, color, xcolor}
\usepackage[pagebackref,breaklinks,colorlinks]{hyperref}
\usepackage{url}
\usepackage{enumerate}
\usepackage {booktabs}
\def\cpar{\hss\egroup\line\bgroup\hss}

\hypersetup{
setpagesize=false,
 bookmarksnumbered=true,%
 bookmarksopen=true,%
 colorlinks=true,%
 linkcolor=blue,
 citecolor=red,
}

\def\red#1{\textcolor{red}{#1}}
\def\blue#1{\textcolor{blue}{#1}}
\def\green#1{\textcolor{green}{#1}}

\title{\LARGE NTIRE 2023 Image Shadow Removal Challenge \\Technical Report: Team IIM\_TTI}

\author{
Yuki Kondo\thanks{Equally contributed author},
Riku Miyata\footnotemark[1],
Fuma Yasue\footnotemark[1],
Taito Naruki\footnotemark[1],
Norimichi Ukita
\\[0.5em]
\large Intelligence Information Media Lab., Toyota Technological Institute (IIM\_TTI)\\
\large \url{{yuki_kondo, ukita}@toyota-ti.ac.jp}
}

\begin{document}

\maketitle

\section*{Abstruct}
In this paper, we analyze and discuss ShadowFormer in preparation for the NTIRE2023 Shadow Removal Challenge~\cite{vasluianu2023ntire}, implementing five key improvements: image alignment, the introduction of a perceptual quality loss function, semi-automatic annotation for shadow detection, joint learning of shadow detection and removal, and the introduction of a new data augmentation technique ``CutShadow'' for shadow removal. Our method achieved scores of 0.196 (3rd out of 19) in LPIPS and 7.44 (4th out of 19) in the Mean Opinion Score (MOS).

\section{Team details}

\begin{itemize}
\item Team name : IIM\_TTI                              
\item Team leader name : Yuki Kondo                          
\item Team leader institution and email (Please make sure is an active email) : Toyota Technological Institute (yuki\_kondo@toyota-ti.ac.jp )
\item Rest of the team members : Riku Miyata (Toyota Technological Institute, sd22431@toyota-ti.ac.jp); Fuma Yasue(Toyota Technological Institute, sd22434@toyota-ti.ac.jp ); Taito Naruki (Toyota Technological Institute,  sd22423@toyota-ti.ac.jp ); Norimichi Ukita (Toyota Technological Institute, ukita@toyota-ti.ac.jp )
        
\item Team website URL (if any) : \url{https://www.toyota-ti.ac.jp/Lab/Denshi/iim/index.html}                
\item Affiliations : Toyota Technological Institute, Japan.
\item Usernames on the NTIRE 2023 Image Shadow Removal Challenge Codalab leaderboard (development/validation and testing phases) : Yuki-11; amaguri; YasueFuma; feln
\item Link to the codes/executables of the solution(s) : \url{https://github.com/Yuki-11/NTIRE2023_ShadowRemoval_IIM_TTI}
\item Link to the enhancement results of all frames : \url{https://drive.google.com/drive/folders/11zu0P7yAORDPhcTBjW5PSYODYi--gMcO?usp=share_link}
\end{itemize}

\section{Contribution details}

\begin{center}
\bfseries\Large
ShadowFormer+ : Rethinking the ShadowFormer learning methods and removing the effect of external camera parameter changes
\end{center}


\vspace{1.5em}

\subsection{Analysis of provided NTIRE 2023 SHADOW REMOVAL dataset}\label{subsection:dataset_analysis}

\begin{figure*}[t]
\centering
    \includegraphics[width=0.8\textwidth]{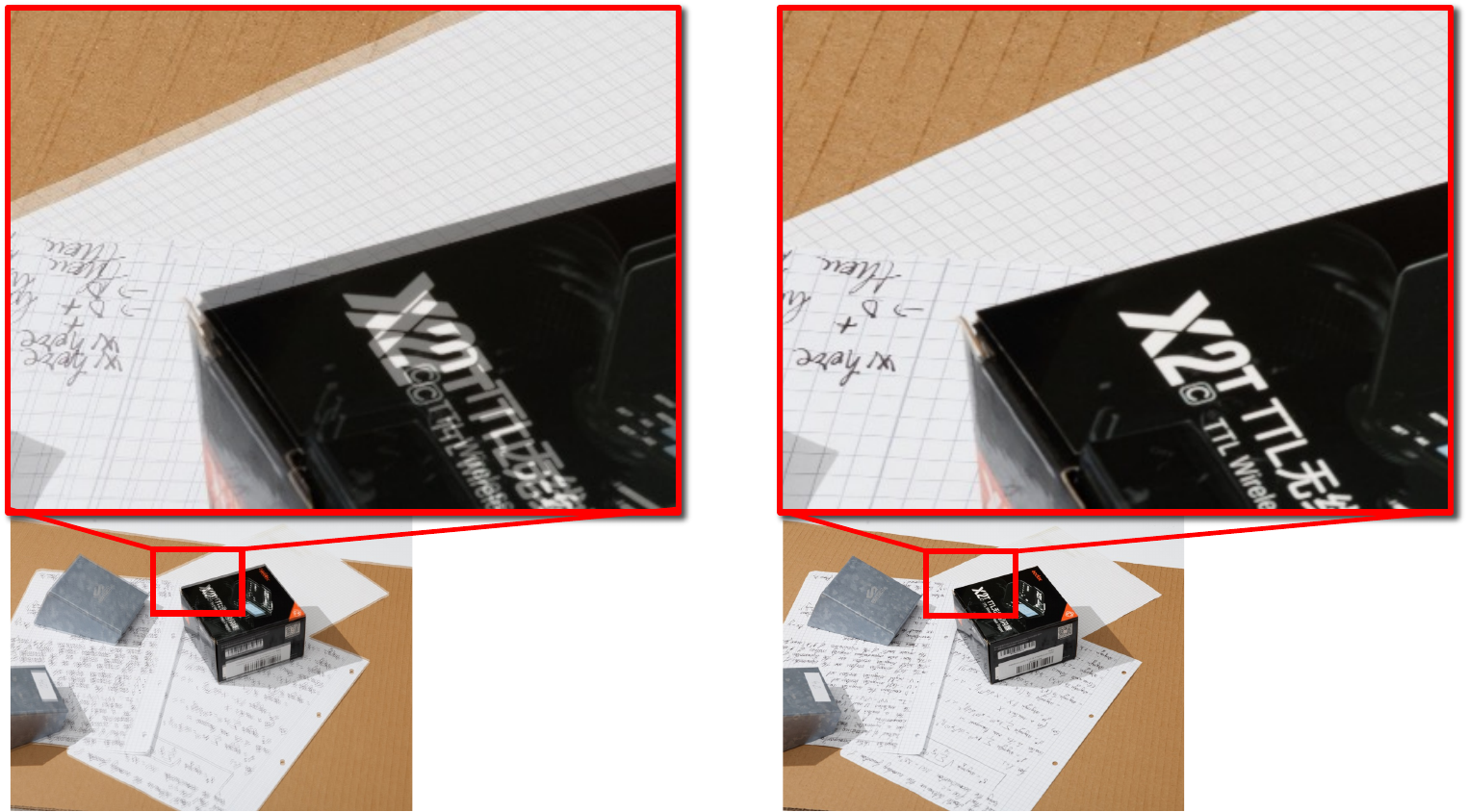}\\
    ~\hspace{-18mm}(a) Misalignment in original sample \hspace{8mm}(b) Aligned sample\\
    
    \caption{Superimposed display of shadow and no-shadow pair images. 
    Random changes in external parameters caused the misalignment (a), but our method was able to correct the misalignment (b).}
\label{fig:warp}
\end{figure*}

After analyzing this challenge's dataset, we found several scenes where the external camera parameters differed significantly between the input image with shadows and its GT image with no shadow, as shown in Fig.~\ref{fig:warp} (a) .
Therefore, in addition to the original task of shadow removal, we have to spatially align these two images by implicitly or explicitly calibrating the external camera parameters between the images.

We also found that many images in the dataset are taken from bird's eye views rather than side views so that a planar regions is dominant in each image.
In addition, such planar regions have local feature points on sheets of papers, clothes, and so on.

Based on the above discussion, we conclude that homography-based image alignment is effective as a preprocess for shadow removal learning.

\subsection{Analysis and discussion of ShadowFormer}
\label{subsection:shadowformer}

ShadowFormer~\cite{shadowformer} is the Transformer-based SOTA model of shadow removal.
ShadowFormer effectively models the context correlation between shadow and non-shadow regions utilizing the Shadow-Interaction Module (SIM) that performs Shadow-Interaction Attention (SIA) in the intermediate feature space.

To further improve ShadowFormer, we found the following five rooms for improvement.

\vspace{1em}
\noindent{\bf (1) Lack of the explicit external camera parameter estimation mechanism:} 
ShadowFormer does not have a function for image alignment between images with and without shadows, which are required in the dataset of this shadow removal challenge.

\vspace{1em}
\noindent{\bf (2) Limitations on perceptual quality improvement of pixel-wise loss functions:} Charbonnier loss is employed in ShadowFormer, but it is known that pixel-wise loss functions in the image space, such as L1 loss and L2 loss, including Charbonnier loss, limit the perceptual quality~\cite{pirm2018, lpips, perceptual_loss, zhang2020ntire}.



\vspace{1em}
\noindent{\bf (3) Difficulty in applying shadow detector to the challenge dataset:}
ShadowFormer requires a shadow mask.
As with the original implementation of ShadowFormer, we use DHAN~\cite{dhan} for detecting the shadow mask in an input image.
Since the dataset of this challenge consists of a wide variety of shadows, the shadow detector should be optimized to the domain of this dataset.
However, since there is no shadow mask in the dataset of this challenge, it is impossible to apply the straightforward supervised learning for optimizing the shadow detector.

\vspace{1em}
\noindent{\bf (4) Independent optimization of shadow detector and shadow remover:}
While it is known that a preprocess (e.g., shadow detector) can be jointly trained with a main task (e.g., shadow remover) for improving the performance of the main task in a variety of tasks~\cite{haris2021task,hayashi2021joint,nakatani2021group,kondo2023joint,kondo2021crack}, ShadowFormer is trained independently with the shadow detector.

\vspace{1em}
\noindent{\bf (5) Insufficient data augmentation:} 
In training of ShadowFormer, only three types of data augmentation schemes are employed: MIxUp~\cite{mixup}, rotation, and vertical flip.
Since the number of training images in the challenge dataset (i.e., 1,000 images), further data augmentation schemes can improve the performance of ShadowFormer.

\vspace{1em}

\subsection{Our method}\label{subsec:our_method}

\begin{figure*}[t]
\centering
    \includegraphics[width=\textwidth]{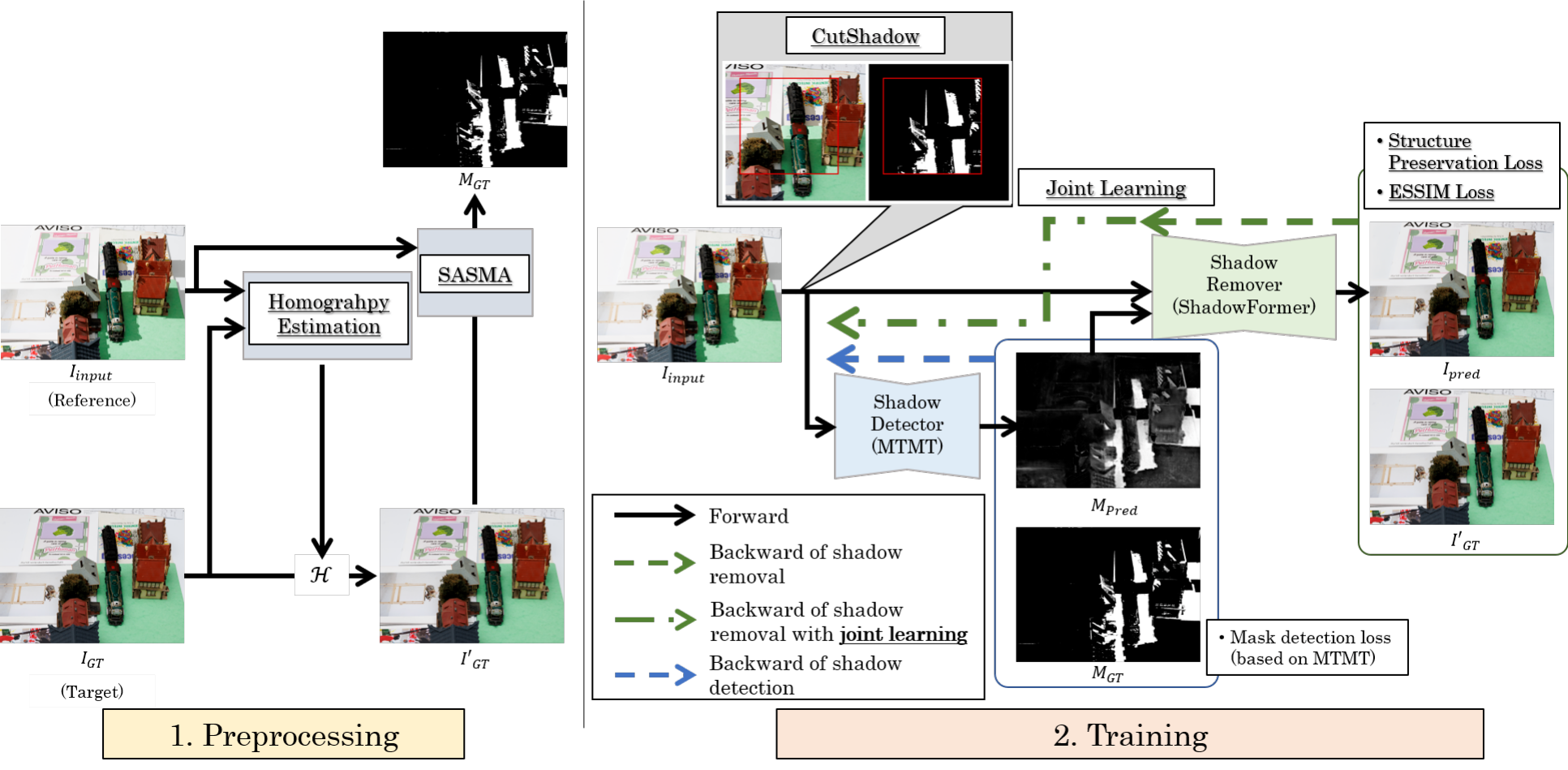}\\
    
    \caption{Overall diagram of our methodology.
    In preprocessing, we correct image misalignment by Homography Estimation, and also create GT shadow mask by SAMSA.
    In training, firstly shadow detector predict the shadow mask from the shadow image. Next,  ShadowFormer predict shadow-free image from shadow image and the shadow mask which predicted by detector.
    at this time, CutShadow is used for augumentation, and ESSIM loss and Structure Preservation Loss are used as the error functions.}
\label{fig:method_overview}
\end{figure*}

In our proposed method, the five problems described in Sec.~\ref{subsection:shadowformer} are resolved as follows.

\vspace{1em}
\noindent{\bf (1) Image alignment by global homography transformation:} 

\begin{figure*}[t]
\centering
    \includegraphics[width=\textwidth]{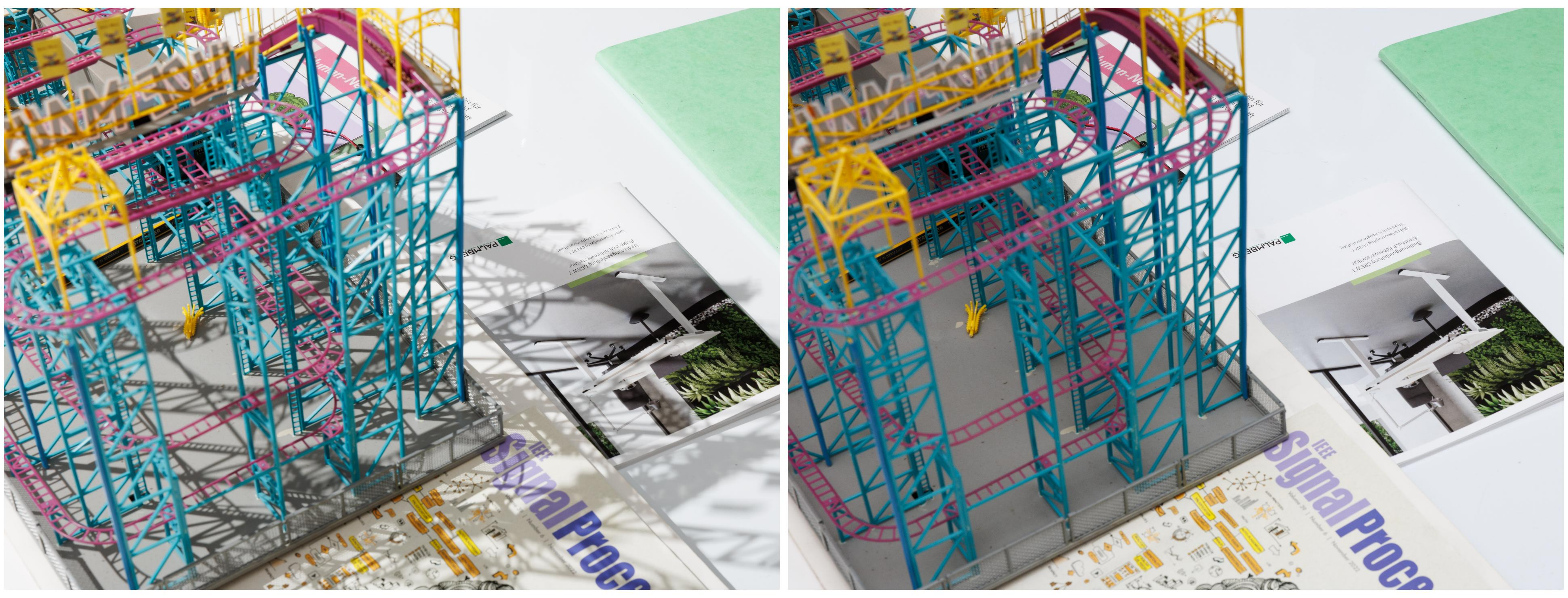}\\
    ~\hspace{10mm}(a) Input image \hspace{25mm}(b) Non-aligned ground truth\\
    \includegraphics[width=\textwidth]{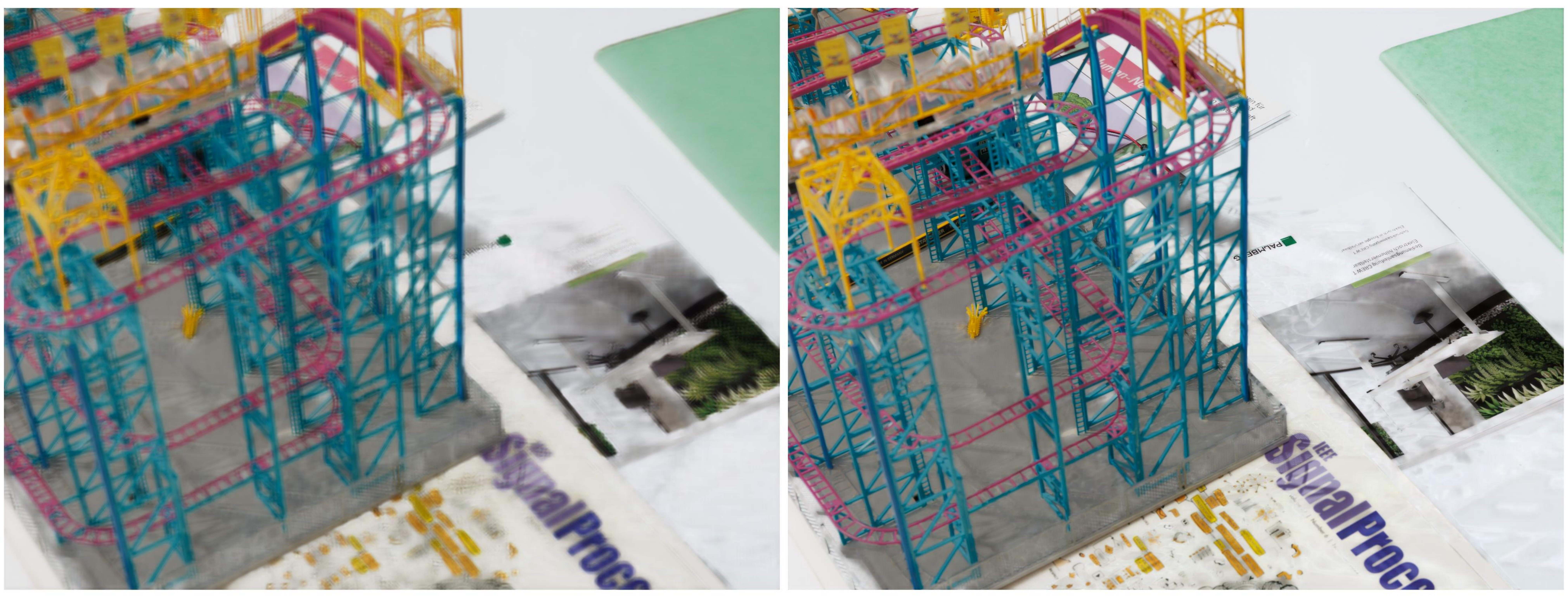}\\
    ~\hspace{2mm}(c) Baseline \hspace{30mm}(d) Ours (Aligned)\\  
    \caption{Comparison of (c) the baseline results of training ShadowFormer without alignment and (d) the results of our method including alignment. The PSNR based on the original unaligned GT in (c) and (d) are 26.79 and 22.67, respectively, and while (c) is superior in PSNR, (d) is perceptually superior because it is not blurred and its structure is guaranteed. Shadows are also accurately removed for the regions where they occur.}
\label{fig:result}
\end{figure*}

As mentioned in Sec.~\ref{subsection:dataset_analysis}, in training, images with and without shadows are spatially aligned based on Homography.
Homography estimation is done with local feature matching using AKAZE~\cite{akaze} and outlier removal using RANSAC~\cite{ransac}.

With the estimated Homography, an image without shadows is transformed to align it with the one with shadows.
Lacking pixels caused by the Homography transformation are the reflect padding process because it improves the shadow removal performance rather than using the zero padding process.
The original training images and their aligned images are shown in Fig.~\ref{fig:warp} (b).
It can be seen that most pixels are successfully aligned.

While we train the shadow remover (i.e., ShadowFormer) with the aligned image pairs, the shadow remover can be trained so that it also implicitly align images with and without shadows if these misaligned images are used as training images.
Therefore, with the misaligned GT images with and without shadows used in the Development and Test phases, the shadow remover trained with such misaligned images works better than the one trained with the aligned images in terms of PSNR.
On the other hand, as shown the result shown in Fig.~\ref{fig:result}, the perceptual quality of the results trained on aligned data is greatly improved.

\vspace{1em}
\noindent{\bf (2-1) Structure Preservation loss:} 
We adopt Structure Preservation loss, which minimizes the distance between the output and GT images in the feature space obtained from DiNO\footnote{Using unlabeled ImageNet~\cite{imagenet} contains about 14 million images for pre-training.}~\cite{dino} in accordance with the success of ShadowDiffusion~\cite{Jin2022ShadowDiffusionDS}, instead of Charbonnier loss.
However, unlike the original Structure Preservation loss, our method does not use the cosine similarity but the euclidean distance using MSE. The reason for this change is that the quantitative results are almost equal between the two similarity measurements, but the visual quality was better with the MSE. The above loss function is shown in the following equation:
\begin{eqnarray}
\mathcal{L}_{SP}&=&\mathrm{MSE}\left(\mathrm{DiNO}\left(I_{pred}\right), \mathrm{DiNO}\left(I'_{GT}\right)\right)\label{eq:sp_loss}
\end{eqnarray}
where DiNO is the DiNO feature extractor, MSE is the mean squared error function, and $I_{pred}$ means the output image from ShadowFormer.

\vspace{1em}
\noindent{\bf (2-2) Edge-based SSIM (ESSIM) loss:}
While Structure Preservation loss is used in the feature space for semantic matching, we also use another loss function in the image domain for preserving the geometric structure.
As validated in~\cite{ssimloss}, it is known that a SSIM-based loss can improve the perceptual quality in contrast to a MSE-based loss.
In the proposed Edge-based SSIM (ESSIM), an input image is first mapped to the HSV space to extract the structure represented in the V-channel map.
Next, the edge image is acquired by Canny edge detector~\cite{canny}, and the SSIM loss is calculated in the edge image:
\begin{eqnarray}
\mathcal{L}_{ESSIM}&=&\mathcal{L}_{SSIM}\left(\mathrm{EdgeDetect}\left(I_{pred}\right), \mathrm{EdgeDetect}\left(I'_{GT}\right)\right),\label{eq:essim_loss}\\
\mathrm{EdgeDetect}(I)&=&\mathrm{Canny}([\mathcal{T}_{HSV}(I)]_{V}),\label{eq:edge_detect}
\end{eqnarray}
where $\mathcal{L}_{SSIM}$ means SSIM loss~\cite{ssimloss}, $\mathcal{T}_{HSV}$ is the HSV translation, $[\ ]_{V}$ means V-channel map extraction and $\mathrm{Canny}$ means Canny edge detection~\cite{canny}.

\begin{figure*}[t]
\centering
    \includegraphics[width=\textwidth]{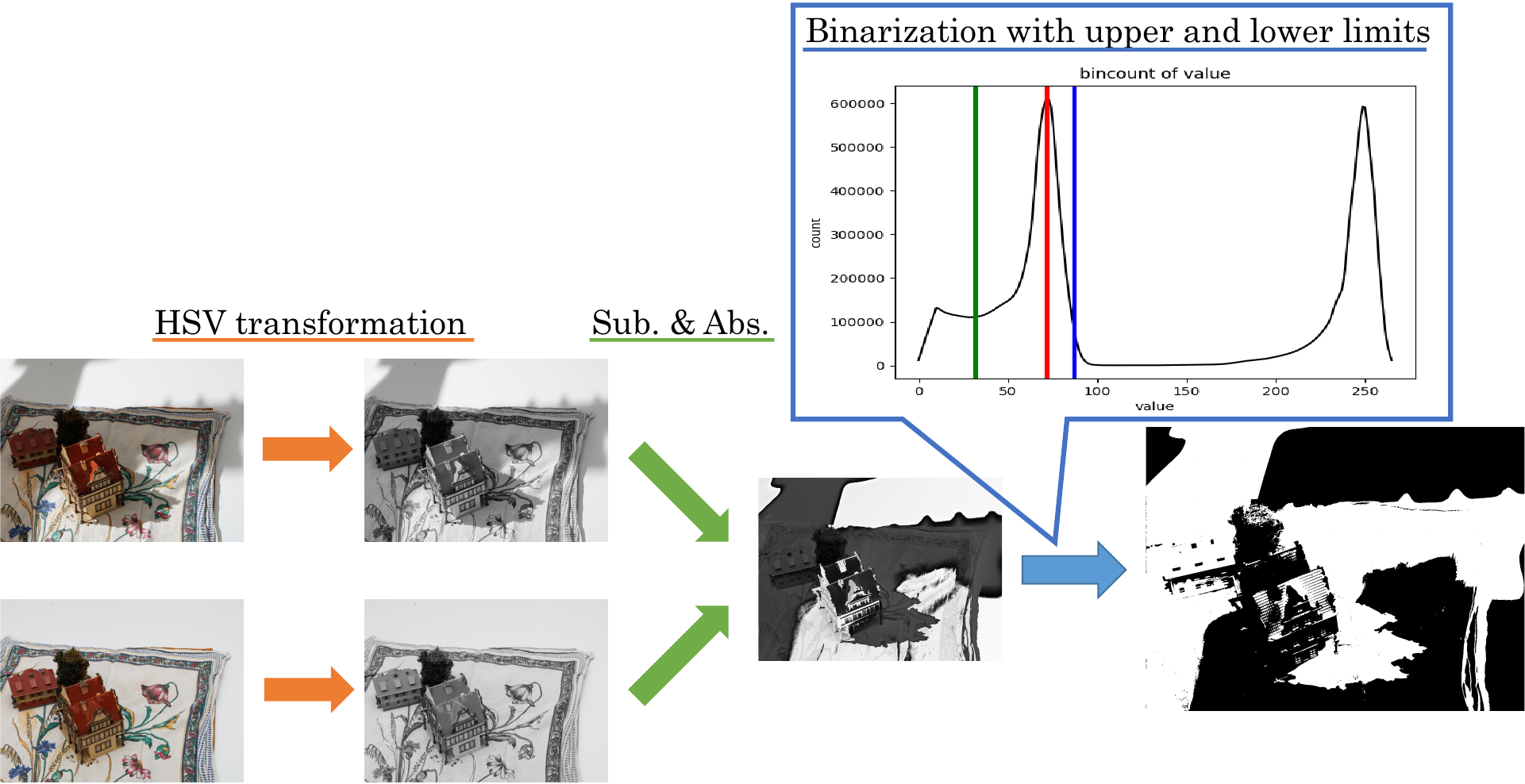}\\
    
    \caption{Semi-Automatic Shadow Mask Annotation (SASMA). The \red{red} line in the histogram indicates the peak value, and the \green{green} and \blue{blue} lines indicate the lower and upper values used for shadow binarization calculated from the peak values, respectively. Note that the second process, Sub.\&Abs., is done in the value channel of HSV.}
\label{fig:mask_gen}
\end{figure*}

\vspace{1em}
\noindent{\bf (3) Semi-Automatic Shadow Mask Annotation:} 
To supervise the shadow detector on the challenge dataset, we need to prepare the GT of a shadow mask. Therefore, we propose Semi-Automatic Shadow Mask Annotation (SASMA), which performs easy, efficient, and reasonable annotation semi-automatically.

SASMA first obtains the absolute error maps of the V channel maps of both the HSV-transformed input and GT images.
The histogram of the error map is displayed to an annotator.
The annotator is requested only to found the distributions corresponding to shadow regions.
With the error values given by the annotator in the histogram, we can determine their corresponding pixels in the image.
These pixels are regarded as shadow pixles.
Figure~\ref{fig:mask_gen} shows the process flow of our SASMA.

\vspace{1em}
\noindent{\bf (4) Joint learning of shadow detector and shadow remover:} 
\begin{figure*}[t]
\centering
    \includegraphics[width=\textwidth]{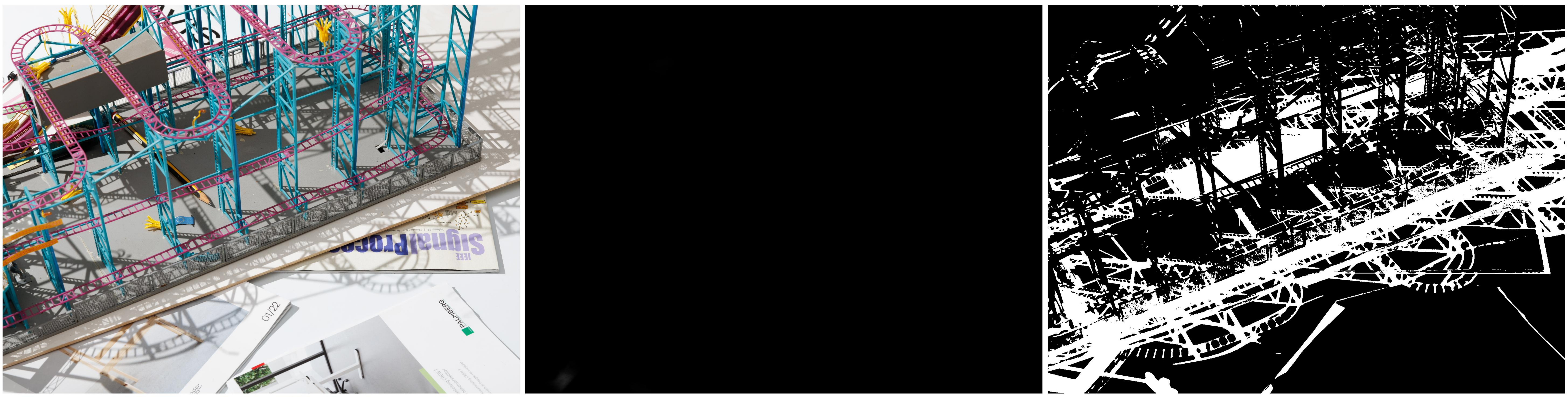}\\
    \scriptsize
    ~\hspace{5mm}(a) Input image \hspace{14mm}(b) MTMT (Released weights)\hspace{4mm}(c) Mask annotated by SASMA\\
    \includegraphics[width=\textwidth]{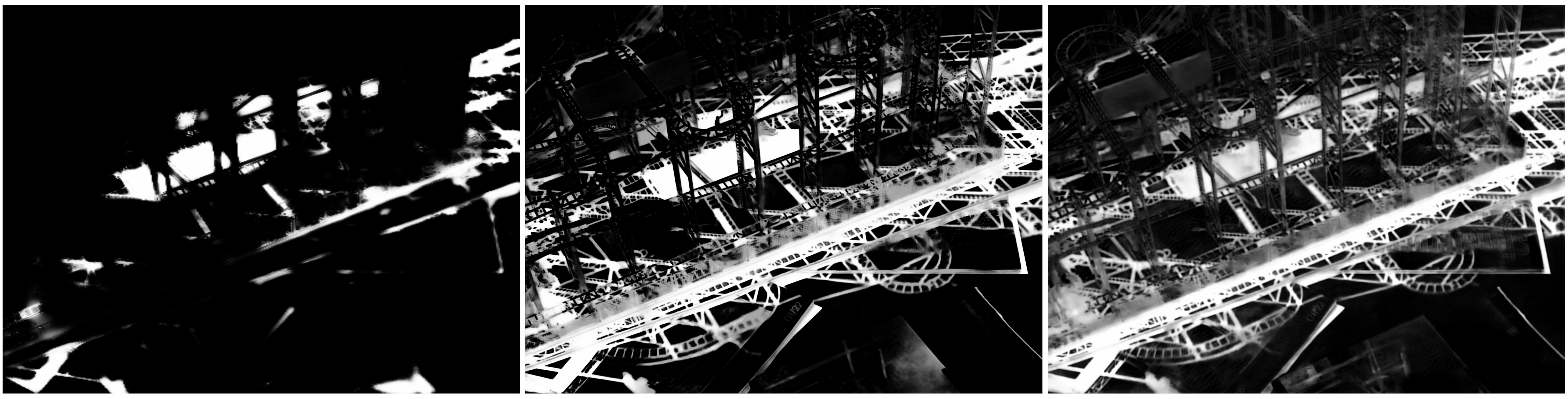}\\
    ~\hspace{2mm}(d) MTMT (Pretiraned) \hspace{7mm}(e) Joint learning ($\alpha=10^{-2}$)\hspace{4mm}(f) Joint learning ($\alpha=10^{-3}$)\\  
    \caption{Refinement of shadow detection by joint learning. At first, MTMT is not able to predict shadows at all (b), but by learning the mask annotated by SASMA (c), it is able to predict shadows to some extent (d). Furthermore, by incorporating joint learning, MTMT is able to predict shadows at the level of the GT mask, and is also able to capture the strength and weakness of the shadows (e) and (f).}
\label{fig:jointlearning_mask}
\end{figure*}

In our joint learning, the shadow detection network is serially connected to the shadow removal network (i.e., ShadowFormer) in order to train these two networks in an end-to-end manner.
For this joint leaning, we use MTMT~\cite{mtmt}, a shadow detection SOTA method, instead of DHAN used in the original ShadowFormer.

The loss function $\mathcal{L}_{joint}$ we use in joint learning is as follows:
\begin{eqnarray}
\mathcal{L}_{joint}&=&(1 - \alpha)\mathcal{L}_{removal}+\alpha\mathcal{L}_{detction},\label{eq:joint}\\
\mathcal{L}_{removal}&=&\beta\mathcal{L}_{SP}+\mathcal{L}_{ESSIM}+\mathcal{L}_{MSE},\label{eq:removal_loss},
\end{eqnarray}
where $\alpha$ is the pretask weight, $\mathcal{L}_{MSE}$ means MES loss, $\beta$ is the weight of the structure preservation loss.
$\beta$ = $10^6$ in our experiments.

The effects of our proposed (3) SASMA and (4) joint learning are shown in Fig.~\ref{fig:jointlearning_mask}.
The result of shadow detection using MTMT with the published weights is shown in Fig.~\ref{fig:jointlearning_mask} (b) in which no shadow is detected.
In our method, on the other hand, a shadow mask annotated by SASMA in each training image, whihc is shown in Fig.~\ref{fig:jointlearning_mask} (c), is used to finetune MTMT.
The result of the finetuned MTMT is shown in Fig.~\ref{fig:jointlearning_mask} (d).
While the performance is significantly improved in (d), many other thin shadows are not detected yet.
On the other hand, shadows detected by our joint learning network, which are shown in Fig.~\ref{fig:jointlearning_mask} (e) and (f), are better than (d).
Interestingly, if $\alpha$ is reduced, i.e., it is dedicated to solving the shadow removal task, the mask becomes sharper, even though the shadow detection task is less important.
This shows that the representation of the mask image required by ShadowFormer is close to that of the GT mask.

\vspace{1em}
\noindent{\bf (5) CutShadow:} 

\begin{figure*}[t]
\centering
    \includegraphics[width=\textwidth]{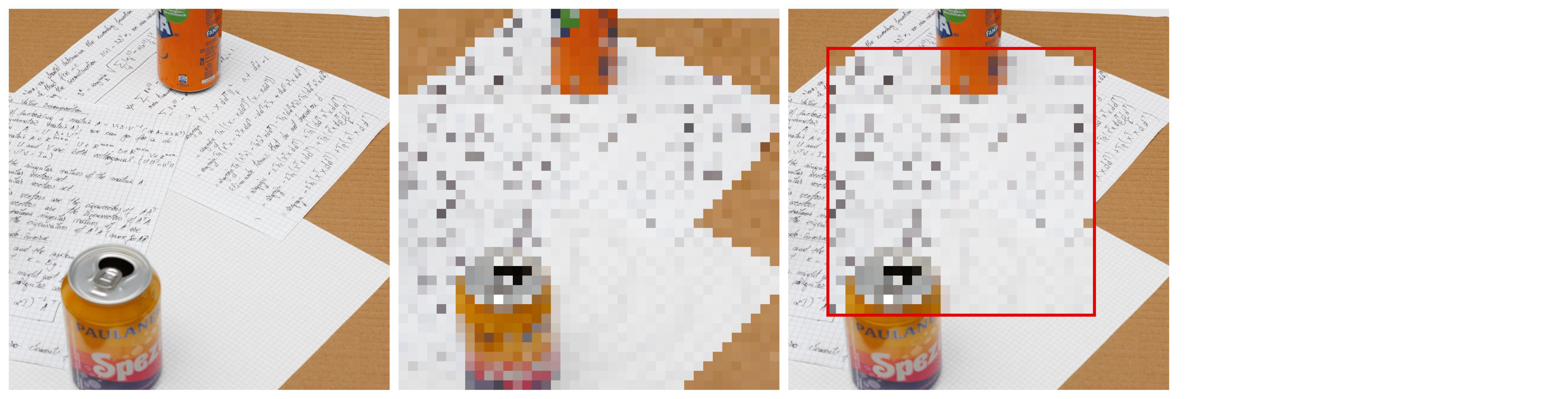}\\
    CutBlur~\cite{yoo2020rethinking}\\
    \includegraphics[width=\textwidth]{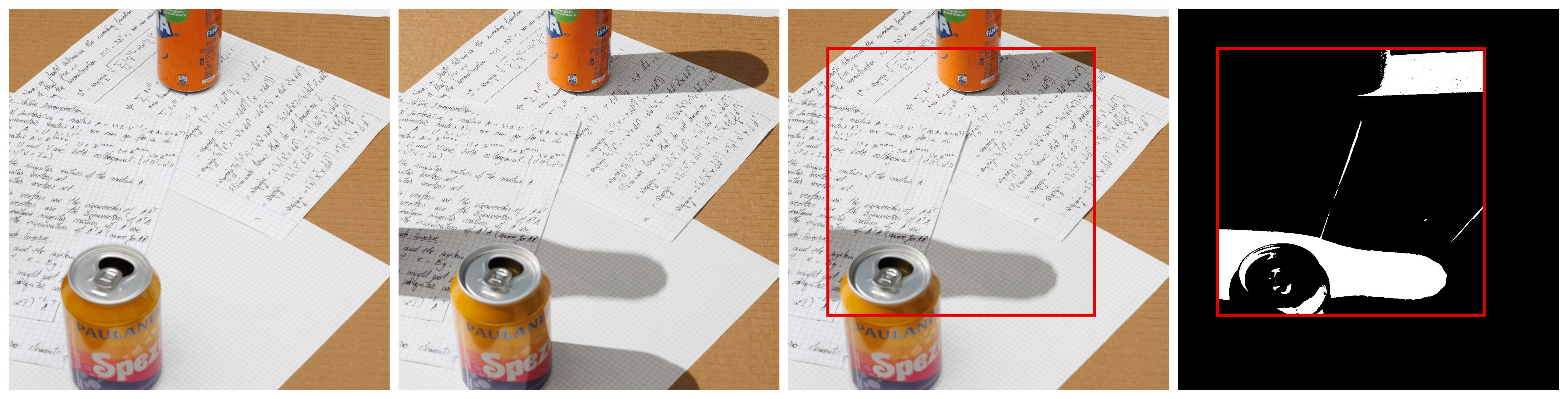}\\
    CutShadow (Ours)\\    
    \caption{Two data augmentations of CutBlur and CutShadow. The first row shows the GT, the second row shows the input image, the third row shows the new input image generated by each data extension (reverse augmentation is also possible, where the base image is the input image and the GT patches are pasted), and the fourth row shows the new input shadow mask generated by CutShadow.}
\label{fig:cutshadow}
\end{figure*}

We propose CutShadow to generate more pseudo-trained data based on CutBlur~\cite{yoo2020rethinking} proposed for image super resolution.
Figure~\ref{fig:cutshadow} shows the samples of CutBlur and CutShadow. 
Given paired training images (e.g., images with and without shadows for the shadow removal task), While low-resolution image patches are pasted on a high-resolution image, image patches with shadows are pasted on an image without shadows.

\section{Global Method Description}

\subsection{Learning Strategies}\label{subsec:lear_stra}

\subsubsection{Preprocessing phase}

\begin{enumerate}
    \item Alignment of GT images by homography transformation
    \item Semi-automatic generation of shadow masks using SASMA
\end{enumerate}

\subsubsection{Training phase}
\begin{enumerate}
    \item Shadow detector (MTMT) pre-training using challenge dataset from initial weights. However, for ResNeXt, which is a module of MTMT, we use the parameters already learned by ImageNet~\cite{imagenet}. 
    \item Shadow remover (ShadowFormer) pre-training using challenge dataset from initial weights.
    \item Joint learning shadow detector and shadow remover using challenge dataset from pre-trained weight.
\end{enumerate}

\subsubsection{Testing phase}
\begin{enumerate}
    \item The provided image is input to the jointed model to obtain a shadow removal image.
\end{enumerate}

\subsection{Quantitative and qualitative advantages of the proposed solution}

\begin{table}[t]
  \centering
  \caption{
  Comparison with baseline and our method.
  The best scores are colored by \textcolor{red}{\textbf{red}} , respectively. Our method is pre-aligned, so the PSNR and SSIM based on the misaligned GT image are extremely low.
  }
  \label{table:result}
  \small
    \begin{tabular}{lcc} 
    \toprule 
    {Model} & {PSNR$\uparrow$} & {SSIM$\uparrow$}\\ 
    \midrule
    Baseline & \textcolor{red}{\textbf{19.92}} & \textcolor{red}{\textbf{0.660}}\\
    \textbf{Ours} & 18.69 & 0.605\\
    \bottomrule
    \end{tabular}
\end{table}

To demonstrate the superiority of the proposed method, we present a qualitative and quantitative evaluation of the plain ShadowFormer for baseline model.

The output samples are shown in the Fig.~\ref{fig:result}. The baseline has a high PSNR evaluated with the original misaligned GT image, but a blurred image is generated by averaging random external camera parameter changes. On the other hand, although the proposed method has a low PSNR evaluated on the original misaligned GT image, it can maintain the context of the scene and remove the shadows, thus obtaining perceptually excellent results.

\subsection{Other Additional Notes}

\vspace{1em}
\noindent{\bf Novelty degree of the solution:}
Please check section~\ref{subsec:our_method}.

\vspace{1em}
\noindent{\bf Previously published:} No.

\vspace{1em}
\noindent{\bf Previously published:}
Please check the table~\ref{tab:tech_info} specifying the technical information.

\begin{table*}[t]
    \centering
    \caption{Technical information table}
    \resizebox{\textwidth}{!}{
    \begin{tabular}{c|c|c|c|c|c|c|c|c}
        Input & Training Time & Epochs & Extra data & Attention & Quantization & \# Params. (M) & Runtime (ms) & GPU  \\
        \hline
          (640, 640, 3)/(1920, 1440, 3) & 60h & 1100 & Yes & Yes & No & 55 Million & 1010 on GPU & A100
    \end{tabular}
    }
    \label{tab:tech_info}
\end{table*}

\section{Competition particularities}
Any particularities of the solution for this competition in comparison to other challenges (if applicable).

\begin{itemize}
\item None
\end{itemize}

\section{Technical details}
Please make sure to write about the language and implementation details: framework, optimizer, learning rate, GPU, datasets used for training, training time, training strategies, efificiency optimization strategies.

\begin{itemize}
\item Framework : PyTorch 1.17
\item optimizer : AdamW
\item learning rate : WarmUp and cosine scheduler (Upper: 0.0002, Lower: 0.000001)
\item GPU : A100
\item datasets used for training : ImageNet~\cite{imagenet} and the challenge dataset with shadow masks.
\item training time : 60 h
\item training strategies : Please check section~\ref{subsec:lear_stra}.
\item efificiency optimization strategies : None
\end{itemize}

\section{Other details}
\begin{itemize}

\item Planned submission of a solution(s) description paper at NTIRE 2023 workshop.
\mbox{}\\
If we win a prize, YES.

\item General comments and impressions of the NTIRE 2023 Image Shadow Removal Challenge (we appreciate your feedback to improve in future editions).\mbox{}\\

It differs from the previous datasets in that the shadow and shadow-free images are misaligned and there is no shadow mask for the shadow images, which was a challenge to devise. These differences are very practical because they are in line with real-life situations.

\item What do you expect from a new challenge in image restoration, enhancement and manipulation?\mbox{}\\

We want to Vison and Language, very sparse video frame interpolation and real world image harmonization challenges.

\end{itemize}

\section*{Acknowledgments}

This work was partly supported by JSPS KAKENHI Grant Numbers 19K12129 and 22H03618.

\bibliographystyle{ieeetr}
\bibliography{ref}

\end{document}